\documentclass[letterpaper, 10 pt, conference]{ieeeconf}  

\IEEEoverridecommandlockouts                              

\usepackage{graphics} 
\usepackage{epsfig} 
\usepackage{times} 
\usepackage{amsmath} 
\usepackage{amssymb}  
\usepackage{bm}
\usepackage{color}
\usepackage{cite}
\usepackage[linesnumbered,ruled]{algorithm2e}
\usepackage{ulem} 
\usepackage[colorlinks=true,linkcolor=black,citecolor=blue,urlcolor=blue,]{hyperref}
\usepackage[doipre={doi:~}]{uri}
\usepackage{float}
\usepackage{booktabs}
\usepackage{multirow}
\usepackage{mathrsfs}
\usepackage[utf8]{inputenc}
\usepackage[caption=false, font=footnotesize]{subfig}
\usepackage{graphicx}
\usepackage{pifont}

\newcounter{RNum}
\renewcommand{\theRNum}{\arabic{RNum}}
\newcommand{\Remark}{\noindent\textit{\textbf{Remark}~\refstepcounter{RNum}\textbf{\theRNum}: }}
\usepackage{soul}
\newcommand{\NoOne}[1]{\textcolor{red}{#1}}
\newcommand{\NoTwo}[1]{\textcolor{green}{#1}}
\newcommand{\NoThree}[1]{\textcolor{blue}{#1}}
 
\soulregister\NoOne7
\soulregister\NoTwo7
\soulregister\NoThree7
\soulregister\Remark7
\usepackage{flushend}

\title{\LARGE \bf
DarkLighter: Light Up the Darkness for UAV Tracking
}

\author{Junjie Ye$^{1}$, Changhong Fu$^{1,*}$, Guangze Zheng$^{1}$, Ziang Cao$^{2}$, and Bowen Li$^{1}$
\thanks{$^{*}$Corresponding author}
\thanks{$^{1}$Junjie Ye, Changhong Fu, Guangze Zheng, and Bowen Li are with the School of Mechanical Engineering, Tongji University, 201804 Shanghai, China.
        {\tt\small changhongfu@tongji.edu.cn}}%
\thanks{$^{2}$Ziang Cao is with the School of Automotive Studies, Tongji University, 201804 Shanghai, China.}
}

\begin{document}

\maketitle
\thispagestyle{empty}
\pagestyle{empty}

\begin{abstract}
Recent years have witnessed the fast evolution and promising performance of the convolutional neural network (CNN)-based trackers, which aim at imitating biological visual systems. However, current CNN-based trackers can hardly generalize well to low-light scenes that are commonly lacked in the existing training set. In indistinguishable night scenarios frequently encountered in unmanned aerial vehicle (UAV) tracking-based applications, the robustness of the state-of-the-art (SOTA) trackers drops significantly. To facilitate aerial tracking in the dark through a general fashion, this work proposes a low-light image enhancer namely DarkLighter, which dedicates to alleviate the impact of poor illumination and noise iteratively. A lightweight map estimation network, \textit{i.e.}, ME-Net, is trained to efficiently estimate illumination maps and noise maps jointly. Experiments are conducted with several SOTA trackers on numerous UAV dark tracking scenes. Exhaustive evaluations demonstrate the reliability and universality of DarkLighter, with high efficiency. Moreover, DarkLighter has further been implemented on a typical UAV system. Real-world tests at night scenes have verified its practicability and dependability.
\end{abstract}

\section{Introduction} \label{sec:intro}
Visual object tracking is a focus in current research of computer vision and has been applied in many unmanned aerial vehicle (UAV)-related applications, \textit{e.g.}, target following~\cite{Cheng2017IROS}, autonomous landing~\cite{Chaudhary2017IROS}, self-localization~\cite{Ye_2021_TIE}, and visual navigation~\cite{Ha_2007_AC}, to name but a few. Motivated by biological neural networks, tracking approaches based on convolutional neural network (CNN) are constructed to track objects like human eyes. Well trained by numerous data, CNN-based trackers promote the tracking performance to the frontier~\cite{LiBo_2019_CVPR, SiamAPN, Fu_2021_TGRS, Bhat_2019_ICCV, Danelljan_2020_CVPR}. However, these trackers can hardly generalize well to low-illumination conditions, since the existing training set commonly lack low-light scenes.


In low-illumination conditions, due to dim light and underexposure, captured images are with high-level noise, low contrast, and low brightness, bringing severe challenges to object tracking. In extreme low-light conditions where naked eyes are hard to distinguish, CNN loses its effectiveness in extracting features. Thus, robust tracking can hardly be maintained without discriminative object features. Evaluations on a dark tracking benchmark demonstrate that even the current top-ranked trackers can hardly keep up their state-of-the-art (SOTA) performance in low-light conditions~\cite{Li_2020_ICRA, 2021arXiv210108446L}. Nevertheless, due to multifarious tracking-related tasks, UAVs inevitably work at night. As a result, the wide application of UAV tracking is hindered by light conditions so far. \textit{How to alleviate the impact of poor illumination on UAV tracking?}
\begin{figure}[!t]	
	\centering
	\includegraphics[width=0.98\linewidth]{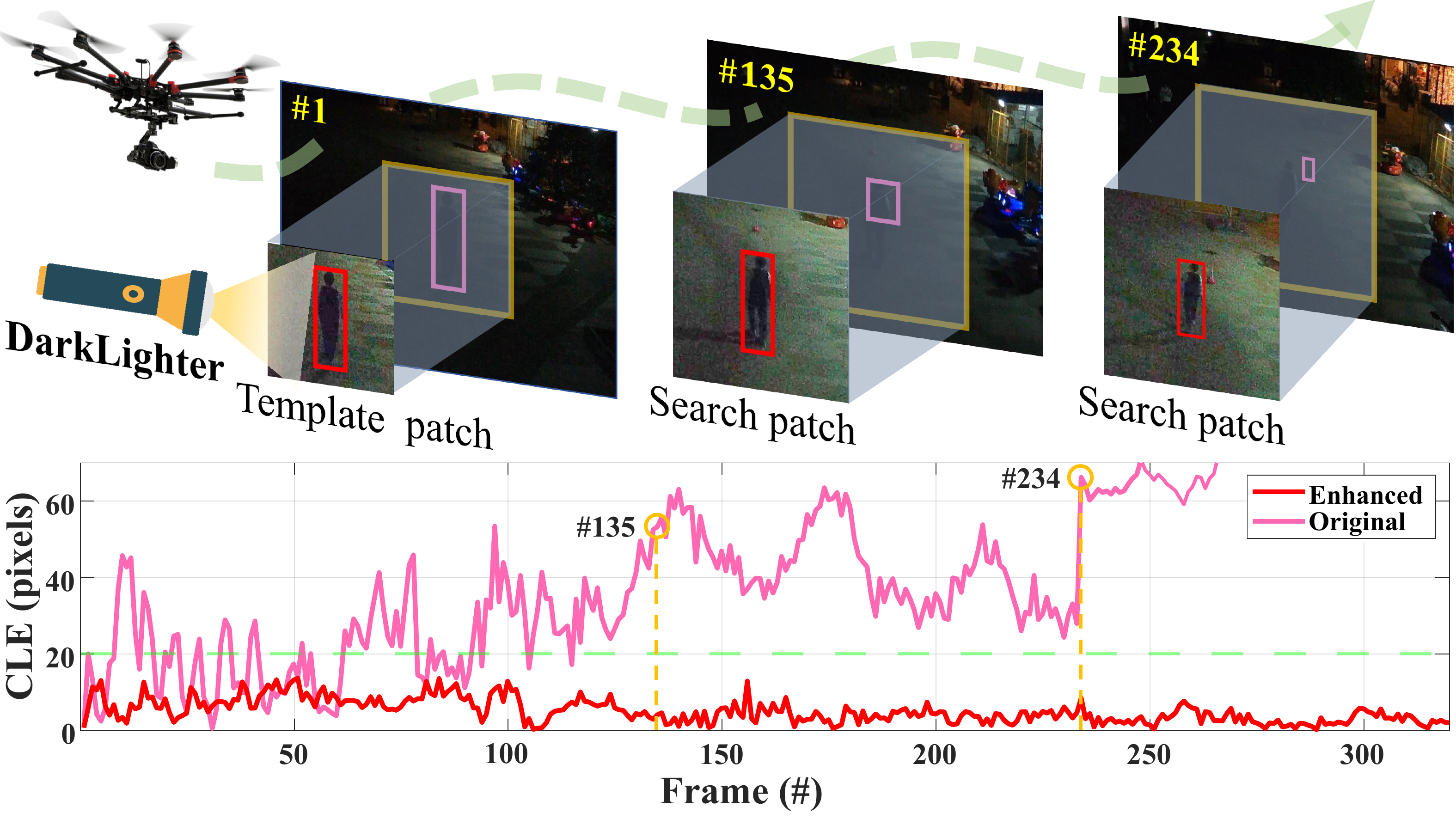}
	\setlength{\abovecaptionskip}{-3pt} 
	\caption
	{
		Tracking performance comparison in a typical dark scene with the proposed DarkLighter activated (in \textcolor{red}{red}) or not (in \textcolor[RGB]{255, 105, 180}{pink}). SiamAPN~\cite{SiamAPN} is adopted as the original tracker here. The center location error (CLE) curves between estimated locations and ground truth bounding boxes are also exhibited. DarkLighter raises tracking robustness in low-light conditions remarkably.
	}
	\label{fig:fig1}
\end{figure}

The drawbacks in dark tracking can be remedied by conducting low-light image enhancement prior to visual tracking, which motivates us to propose a plug-and-play enhancer for UAV tracking. In consideration of the limited computing resources and complex working conditions of UAVs, such an enhancer must be lightweight, robust, easy to deploy, and most importantly---beneficial to object tracking.

Towards adjusting the illumination and improving the visual quality of dark images, low-light enhancement has attracted widespread attention. The target of most previous works on low-light enhancement is mainly promoting aesthetic quality. Besides, the main metrics in image enhancement, \textit{e.g.}, peak signal to noise ratio (PSNR), and structural similarity (SSIM), are designed to evaluate signal fidelity. However, low-light enhancement for high-level tasks, for instance, object tracking, has received little attention so far.

\begin{figure*}[!t]	
	\includegraphics[width=0.98\linewidth]{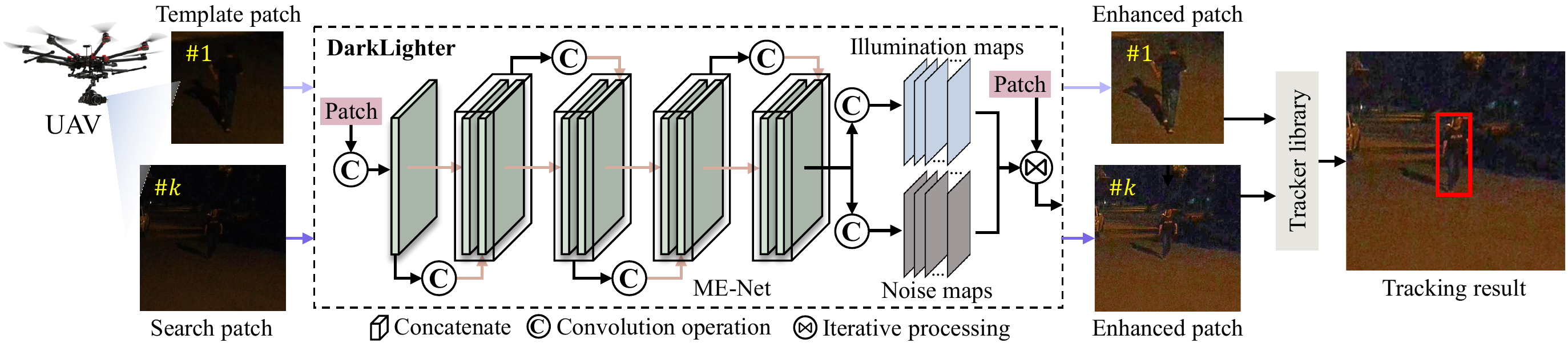}
	\setlength{\abovecaptionskip}{-2pt} 
	\caption
	{
		The pipeline of performing low-light enhancement prior to object tracking. Given a low-light image patch, DarkLighter dedicates to remove the influence of poor illumination processively, lighting up objects for trackers with slight computational consumption.
	}
	\label{fig:main}
\end{figure*}

According to Retinex theory~\cite{Edwin_1977_retinex}, an observed image can be decomposed to reflectance and illumination maps. The reflectance, \textit{i.e.}, the wavelength of the light reflected by the object, is supposed to remain constancy since it is determined by the properties of the object. Therefore, reflectance can be regarded as an object's "true color". Naturally thinking, tracking the object with its "true color" regardless of the light conditions leads to more robust performance. To this purpose, a novel Retinex-inspired image enhancer namely DarkLighter is proposed to facilitate UAV tracking. Specifically, DarkLighter iteratively decomposes the reflectance map from low-light images. A lightweight \textbf{m}ap \textbf{e}stimation \textbf{net}work (ME-Net) is constructed to estimate illumination maps and noise maps for the iteration process of DarkLighter. Adopting the promising Siamese-based trackers as baselines, DarkLighter is integrated into object tracking as a preprocessing step, dedicating to light up the template and search patches before they are fed into the backbone. Since objects commonly appear in the center region of the patches according to the workflow of object tracking, DarkLighter pays more attention to the central area. As shown in Fig.~\ref{fig:fig1}, the activation of DarkLighter raises the original tracker's robustness in the low-light condition. Figure~\ref{fig:main} exhibits the pipeline of tracking framework appending DarkLighter.

The main contributions are summarized below:
\begin{itemize}
	\item A novel Retinex-inspired image enhancer, \textit{i.e.}, DarkLighter, is proposed to iteratively alleviate the influence of poor illumination on UAV tracking. 
	\item  A novel lightweight ME-Net is constructed to estimate illumination and noise maps jointly. No need for paired images, ME-Net is trained with general captured images and inferences efficiently. 
	\item Experiments on numerous UAV dark tracking scenes demonstrate that DarkLighter indeed improves the tracking performance of baselines significantly.
	\item DarkLighter is further implemented on a typical UAV system. Onboard tests corroborate its practicability and dependability in real-world night conditions.
\end{itemize}

\section{Related Works}
\subsection{CNN for UAV Tracking}
Generally, object tracking approaches can be categorized as correlation filter-based approaches~\cite{Danelljan2017CVPR, Fu_2020_TGRS_YJJ,zheng2021ICRA} and CNN-based approaches~\cite{LiBo_2019_CVPR, Xu_2020_AAAI, Cao2021ICCV}. On account of the appealing tracking performance, CNN-based trackers become the current trends of visual tracking. Among them, Siamese-based approaches~\cite{Bertinetto_2016_ECCV, Tao_2016_CVPR, Li_2018_CVPR, LiBo_2019_CVPR, Xu_2020_AAAI, SiamAPN} stand out due to their well-balance of accuracy and efficiency, which are considered to be a solid choice for UAV tracking.
Generally, Siamese-based approaches can be categorized as anchor-based trackers and anchor-free trackers. The anchor-based trackers~\cite{Li_2018_CVPR,LiBo_2019_CVPR} draw lessons from object detection and introduce the region proposal network (RPN)~\cite{Ren2017tpami} to generate precise proposals, achieving remarkable tracking accuracy. To avoid the hyper-parameters accompanying with RPN, anchor-free trackers~\cite{Xu_2020_AAAI} regress the offsets of the bounding box directly. In addition, SiamAPN~\cite{SiamAPN} designs an anchor proposal network for adaptive anchor proposing.
Recently, an online learning strategy is introduced in~\cite{Bhat_2019_ICCV, Danelljan_2020_CVPR} to address the poor generalization of Siamese trackers and make full use of background appearance information. 
However, despite the fantastic performance of these trackers in normal-illumination conditions, their robustness drops greatly in low-light conditions. How can we cast a light on low-light images for the trackers to see?
\subsection{Approaches for Low-light Image Enhancement}
To answer the question, low-light enhancement inspires new vitality for UAV dark tracking. The assumption of Retinex theory~\cite{Edwin_1977_retinex} has motivated a variety of approaches to accomplish decomposition work for the reflectance of the image~\cite{Ren2020TIP, Park2017TCE}. LIME~\cite{LIME} estimates the illumination through the maximum value in RGB channels and refined the illumination map. M. Li \textit{et al.}~\cite{Li_2018_robust_retinex} additionally take noise map into consideration and propose the robust Retinex model. 
The combination of Retinex theory with CNNs has also achieved remarkable performance~\cite{RetinexNet, Zhang2019ACM}. Notably, lightweight methods are also emerging. EnlightenGAN~\cite{EnlightenGAN} is an unsupervised method, which can be trained on unpaired low/normal light data. Zero-DCE~\cite{Guo_2020_CVPR} employs image-specific curve estimation to accomplish image enhancement tasks.
However, tracking onboard UAV requires approaches for both high processing speed and suitability for trackers. The dissatisfaction of real-time speed strictly constrained the application of non-lightweight methods on UAVs. For those lightweight enhancers, they are generally proposed for the improvement of visual quality instead of tracking tasks. Therefore, constructing a new enhancer for facilitating tracking robustness is urgent.

\section{Methodology}
\subsection{Proposed DarkLighter}
\subsubsection{Main Idea of DarkLighter}
Based on the classic Retinex theory~\cite{Edwin_1977_retinex}, the observed image $\mathbf{S} \in \mathbb{R}^{w \times h \times 3}$ can be separated into a reflectance component $\mathbf{R} \in \mathbb{R}^{w \times h \times 3}$ and an illumination component $\mathbf{L} \in \mathbb{R}^{w \times h}$:
\begin{equation}\label{equ:Retinex} 
	\mathbf{S}= \mathbf{R} \odot \mathbf{L}
	\quad,
\end{equation}
where $\odot$ denotes pixel-wise multiplication. $w$ and $h$ are the width and height of the image. The illumination map~$\mathbf{L}$ bounds up with the lightness of the environment. The reflectance map $\mathbf{R}$ is determined by the physical characteristics of objects, which is independent of illumination.

\Remark Since the reflectance relies only on the properties of the object and is regarded as the object's "true color", this work dedicates to feed the tracking backbone with the reflectance of patches, so as to alleviate the impact of poor illumination on UAV tracking. 

Taking intensive noise in low-light images into consideration, a noise term $\mathbf{N} \in \mathbb{R}^{w \times h}$ is added to the classic Retinex model~\cite{Li_2018_robust_retinex}:
\begin{equation}\label{equ:RoRetinex} 
	\mathbf{S}= \mathbf{R} \odot \mathbf{L} + \mathbf{N}
	\quad.
\end{equation}

Once $\mathbf{L}$ and  $\mathbf{N}$ are estimated, the reflectance $\mathbf{R}$ can be decomposed out from the input image as:
\begin{equation}\label{equ:R} 
	\mathbf{R} = (\mathbf{S}-\mathbf{N}) \oslash \mathbf{L}
	\quad,
\end{equation}
where $\oslash$ represents pixel-wise division. 

Assuming observed scenes' illumination and noise maps in a normal lighting condition are an all-ones matrix and a zero matrix, respectively, the obtained reflectance map $\mathbf{R}$ can be regarded as the output of Retinex-based approaches. Since Eq.~(\ref{equ:R}) is inherently an ill-posed problem, directly decomposing the input images may lead to unrealistic results. We propose a Retinex-based low-light enhancer, \textit{i.e.}, DarkLighter, to peel off illumination and noise maps iteratively:
\begin{equation}\label{equ:S} 
	\mathbf{S}_i = (\mathbf{S}_{i-1} - \mathbf{N}_i) \oslash \mathbf{L}_i
	\quad,
\end{equation}
where $i$ denotes the $i$-th iteration. Note that $\mathbf{S}_0$ denotes the initial image and $\mathbf{S}_i$ ($i>0$) indicates the intermediate result. To avoid the situation that the dividend is zero, Eq.~(\ref{equ:S}) is reformulated as:
\begin{equation}\label{equ:S_1} 
	\mathbf{S}_i = (\mathbf{S}_{i-1} - \mathbf{N}_i) \odot \mathbf{E}_i
	\quad,
\end{equation}
where $\mathbf{E}_i = \mathbf{1} \oslash \mathbf{L}_i$. Given a low-light image $\mathbf{S}_0$, DarkLighter erases illumination and noise progressively. After a total of $I$ iterations, the illumination and noise maps are relatively cleaned up, and the final output $\mathbf{S}_I$ is regarded as the reflectance $\mathbf{R}$ of the image (we employ $I = 8$ in this work).

\subsubsection{Outline of ME-Net} 
\begin{figure}[!t]	
	\centering
	\includegraphics[width=0.48\textwidth]{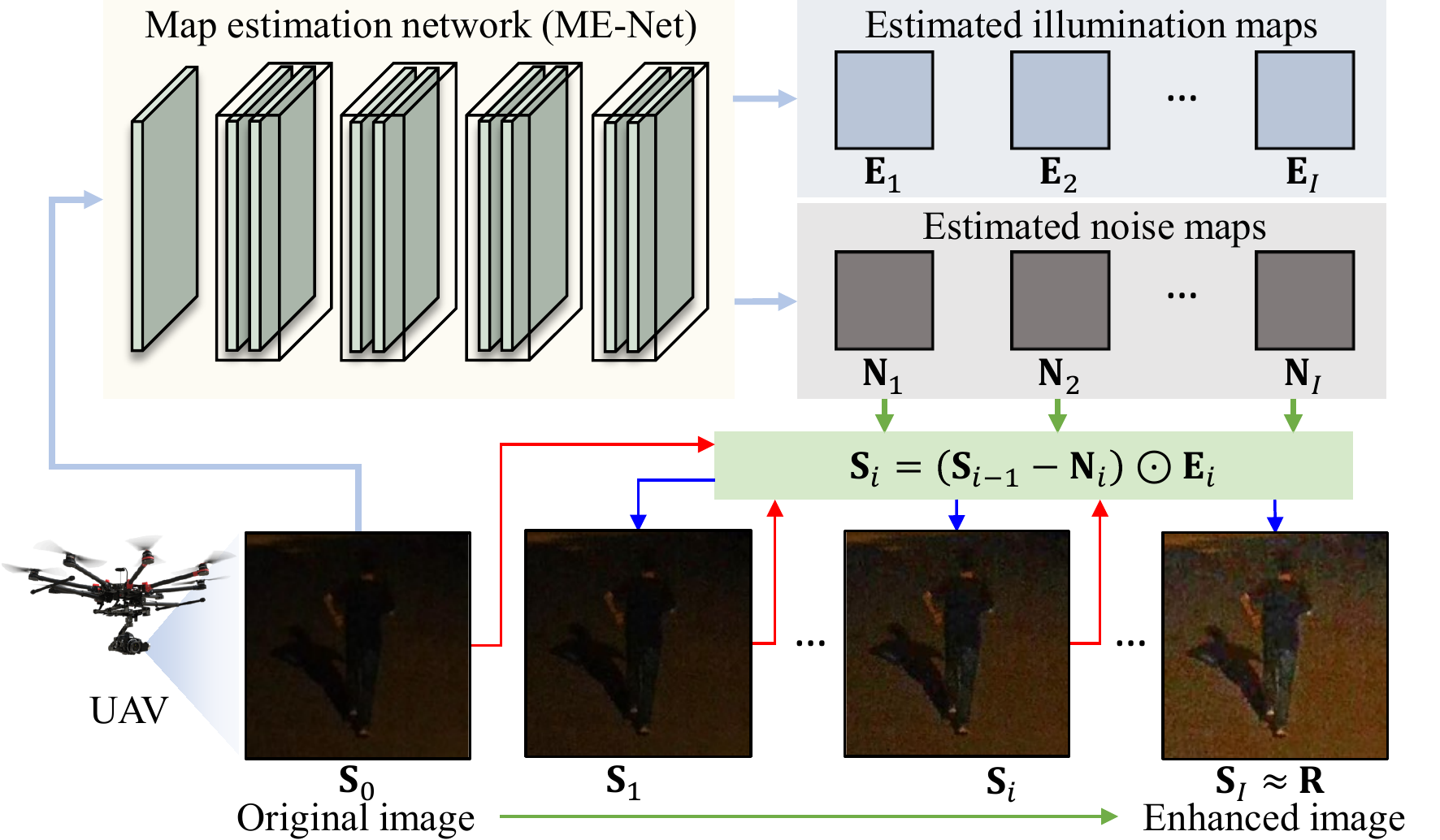}
	\setlength{\abovecaptionskip}{-15pt} 
	
	\caption
	{
		The workflow of the proposed DarkLighter. The reflectance of patches is extracted progressively.
	}
	\label{fig:model}
	\vspace{-0.4cm}
\end{figure}

Recall that $\mathbf{N}$ and $\mathbf{E}$ need to be estimated for every iteration in Eq.~(\ref{equ:S_1}), we train the ME-Net to estimate image-specific illumination and noise maps. Specifically, ME-Net consists of seven convolutional layers as shown in Fig.~\ref{fig:model}. Taking the initial low-light image as input, after passing through several convolutional layers, a set of illumination maps and noise maps are obtained for iterative decomposition. TABLE~\ref{tab:LENET} displays the specific settings of each map in ME-Net. Once $[\mathbf{E}_1, \mathbf{E}_2, ..., \mathbf{E}_I]$ and $[\mathbf{N}_1, \mathbf{N}_2, ..., \mathbf{N}_I]$ are obtained, the reflectance $\mathbf{R}$ (\textit{i.e.}, $\mathbf{S}_I$) of the object is calculated following Eq.~(\ref{equ:S_1}). 
\begin{table}[!b]
	\centering
	\scriptsize 
	\caption{Detailed settings of ME-Net.}
	\renewcommand\tabcolsep{3.pt}
	\resizebox{0.94\linewidth}{!}{
    \begin{tabular}{cccccc}
	\toprule
	Input  & Data dimensions & Operator & Kernel & Stride & Output \\
	\midrule
	$\mathbf{S}_0$    & 256$\times$256$\times$3 & Conv\&ReLU & 3$\times$3 & 1      & Conv1 \\
	Conv1  & 256$\times$256$\times$32 & Conv\&ReLU & 3$\times$3 & 1      & Conv2 \\
	Conv1\&Conv2 & 256$\times$256$\times$64 & Conv\&ReLU & 3$\times$3 & 1      & Conv3 \\
	Conv2\&Conv3 & 256$\times$256$\times$64 & Conv\&ReLU & 3$\times$3 & 1      & Conv4 \\
	Conv3\&Conv4 & 256$\times$256$\times$64 & Conv\&ReLU & 3$\times$3 & 1      & Conv5 \\
	Conv4\&Conv5 & 256$\times$256$\times$64 & Conv\&Tanh & 3$\times$3 & 1      & $\mathbf{L}$ \\
	Conv4\&Conv5 & 256$\times$256$\times$64 & Conv\&Tanh & 3$\times$3 & 1      & $\mathbf{E}$ \\
	$\mathbf{L}$      & 256$\times$256$\times$8 & Split  & -      & -      & $[\mathbf{L}_1, \mathbf{L}_2, ..., \mathbf{L}_I]$ \\
	$\mathbf{E}$      & 256$\times$256$\times$8 & Split  & -      & -      & $[\mathbf{E}_1, \mathbf{E}_2, ..., \mathbf{E}_I]$ \\
	\bottomrule
\end{tabular}}
\label{tab:LENET}%
\end{table}%

\subsubsection{Loss Functions}
To enable ME-Net with robust maps estimation ability, several consumption functions are designed to guide the training process. 

\noindent \textbf{Center-focused lightness loss.} DarkLighter towards lighting up the template patch and the search patch. Since the tracked object is centered at the template patch and generally appears in the central area of the search patch, it is worth paying more attention to enhance the central area. \cite{Guo_2020_CVPR} constraints the average intensity of local regions to an appropriate level, inspired by which, we design a spatial weight map $\mathbf{W}\in \mathbb{R}^{p,q}$ to attract the attention of the enhancer to the central area. Specifically, the enhanced image is divided into $P$ nonoverlapping patches with the size of $16\times16$, the average intensity value $y$ of each patch is regularized to approximate a well-illumination value~$l$. The values in $\mathbf{W}$ is calculated as $w = \ln(e+\sqrt{j^2 + k^2})$, where $j$ and $k$ is assigned according to the spatial location of the corresponding patch, and $e$ is Euler constant. On this basis, the center-focused lightness loss is formulated as:
\begin{equation}\label{equ:lightLoss} 
	\mathcal{L}_{\rm cen} = \frac{1}{P} \left\Vert \mathbf{W} (\mathbf{Y} - l \cdot \mathbf{1}) \right\Vert_{2}^2
	\quad,
\end{equation}
where $\mathbf{Y}$ is a matrix composed of $y$ of each patch arranged in corresponding positions. The well-illumination value $l$ is set to 0.6 in implementation.

\noindent\textbf{Illumination regularization loss.}
For the sake of ensuring that the values of neighboring pixels change monotonically between iterations. The illumination regularization loss is added to constrain the smoothness of each illumination component $\mathbf{E}_i$, which is defined as:
\begin{equation}\label{equ:illLoss} 
	\mathcal{L}_{\rm ill} = \frac{1}{I} \sum_{i=1}^{I}\left\Vert \nabla \mathbf{E}_i \right\Vert_{2}^2
	\quad,
\end{equation}
where $\nabla$ is the first-order differential operator.

\noindent\textbf{Noise estimation loss.} High-level noise inevitably appears in low-light captured sequences. Considering that CNNs already suffer from extracting discriminative features in low-light conditions, such high-level noise can mislead trackers easily. Therefore, it is essential to guide the mapping between image and noise maps properly. A noise estimation loss in DarkLighter is added to restrain the overall intensity of noise maps $\mathbf{N}_i$, which is formulated as:
\begin{equation}\label{equ:noiLoss} 
	\mathcal{L}_{\rm noi} = \frac{1}{I} \sum_{i=1}^{I}\left\Vert \mathbf{N}_i \right\Vert_{2}^2
	\quad.
\end{equation}

\noindent\textbf{Color intensity loss.} Taking the intensity of each channel into consideration, the color intensity loss is designed to bridge three channels and constrain their intensity to a relatively same level. To this concern, the color intensity loss is constructed as:
\begin{equation}\label{equ:colorLoss} 
	\begin{split}
	&\mathcal{L}_{\rm col} = \frac{1}{w\times h}\sum_{(m,n)} \left\Vert \mathbf{R}^m - \mathbf{R}^n \right\Vert_{2}^2\\ 
	&s.t., \quad (m,n) \in \{(r,g),(r,b),(g,b)\}
	\end{split}
	\quad,
\end{equation}
where $\mathbf{R}^*$ denotes the $*$-th color layer of $\mathbf{R}$.

\noindent \textbf{Semantic fidelity loss.}
Since the semantic information of images plays a pivotal role in object tracking, the degradation of semantic information in the enhanced image can easily lead to tracking failure. To this end, a semantic fidelity loss is introduced to the overall loss functions. Denoting $\mathcal{F}(\cdot)$ as the feature extraction operation, the semantic fidelity loss is defined as:
\begin{equation}\label{equ:semLoss} 
	\mathcal{L}_{\rm sem} = \left\Vert \mathcal{F}(\mathbf{R})-\mathcal{F}(\mathbf{S}_0) \right\Vert_{2}^2
	\quad.
\end{equation}

In practice, VGG-16~\cite{Karen_2015} is used to measure the semantic fidelity loss.

Put it all together, the overall loss function is constructed as follows:
\begin{equation}\label{equ:overall} 
	\begin{split}
		\mathcal{L} =\lambda_1 \mathcal{L}_{\rm col} + \lambda_2 \mathcal{L}_{\rm cen} + \lambda_3 \mathcal{L}_{\rm ill} + \lambda_4 \mathcal{L}_{\rm sem} + \lambda_5 \mathcal{L}_{\rm noi}
	\end{split}
	\quad,
\end{equation}
where $\lambda_1$, $\lambda_2$, $\lambda_3$, $\lambda_4$, and $\lambda_5$ are coefficients to adjust the contributions of each consumption function.
\subsection{DarkLighter for UAV Tracking}
\begin{figure}[!t]	
	\centering
	\includegraphics[width=0.98\linewidth]{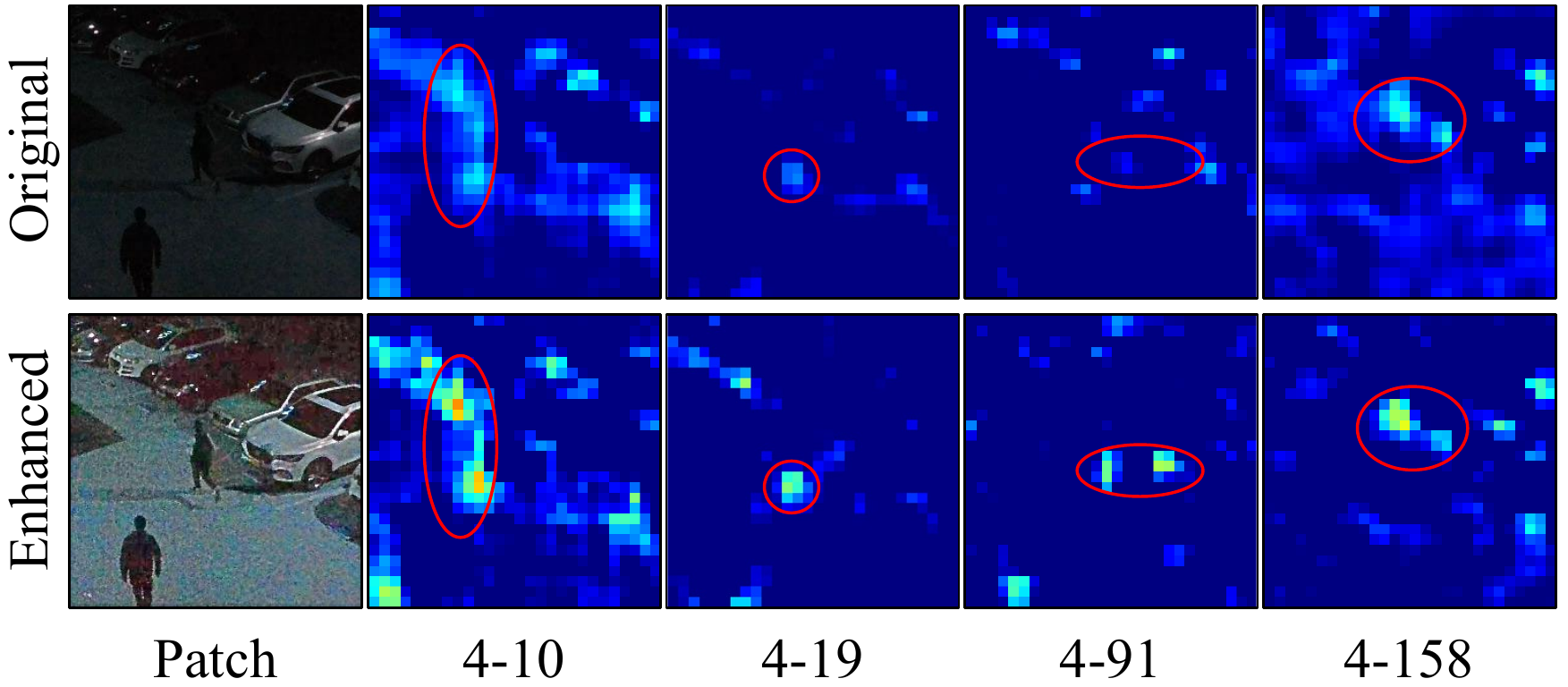}
	\setlength{\abovecaptionskip}{-2.5pt} 
	
	\caption
	{
		Visualization of search patch and features extracted by the backbone. The first row is patch and features from the original image, while the second row is patch lighted by DarkLighter and its corresponding features. "4-10" represents the output of the 10-th channel from the 4-th layer of the backbone. It can be seen clearly that features extracted from the enhanced patch are more salient and discriminable compared to those from the original patch.
	}
	\label{fig:comfeat}
	\vspace{-0.4cm}
\end{figure}

As illustrated in Fig.~\ref{fig:main}, DarkLighter is integrated as a preprocessing module in a CNN-based tracker. The template patch and search patch are lighted up by DarkLighter before being fed into the feature extraction network. A qualitative comparison of features extracted from the original patch and the enhanced one is shown in Fig.~\ref{fig:comfeat}. The features extracted from the original patch seem messy and indistinguishable, while those from the enhanced patch are salient and discriminable. Therefore, the dark tracking performance of trackers is raised up by more discriminative features.

Moreover, it seems appealing to co-train DarkLighter and CNN-based trackers, with the objective of promoting tracking accuracy. However, such an operation inevitably introduces a specific tracking model, degrading the generalization of DarkLighter in other trackers. In this case, the implementation of DarkLighter on any new tracking pipeline needs additional training to guarantee the gain on tracking accuracy. Thus, with the purpose of constructing a general enhancer for UAV tracking, the training of DarkLighter is implemented independently of the tracking baseline. Benefiting from this, DarkLighter is plug-and-play. Without any extra tuning, it is compatible with any CNN-based trackers.

\section{Experiment and Practice}

To testify the effectiveness and robustness of DarkLighter, it is implemented on 4 SOTA trackers, \textit{i.e.}, DiMP50~\cite{Bhat_2019_ICCV}, PrDiMP50~\cite{Danelljan_2020_CVPR}, SiamAPN~\cite{SiamAPN}, and SiamRPN++~\cite{LiBo_2019_CVPR}. Comprehensive experiments are conducted on UAVDark135~\cite{2021arXiv210108446L}---the only UAV dark tracking benchmark so far. In UAVDark135, there are 135 challenging sequences of various objects (person, car, building, \textit{etc.}) captured in poor illumination, with over 125K frames in total, where most dark tracking scenes are involved. 
Moreover, DarkLighter is further compared with SOTA low-light enhancers, respectively EnlightenGAN~\cite{EnlightenGAN}, Zero-DCE~\cite{Guo_2020_CVPR}, and LIME~\cite{LIME}, to demonstrate its advantages for UAV dark tracking. In the end, DarkLighter is further ported to a typical UAV platform to evaluate its applicability and efficiency.

\begin{figure*}[!t]	
	\centering
	\includegraphics[width=0.49\linewidth]{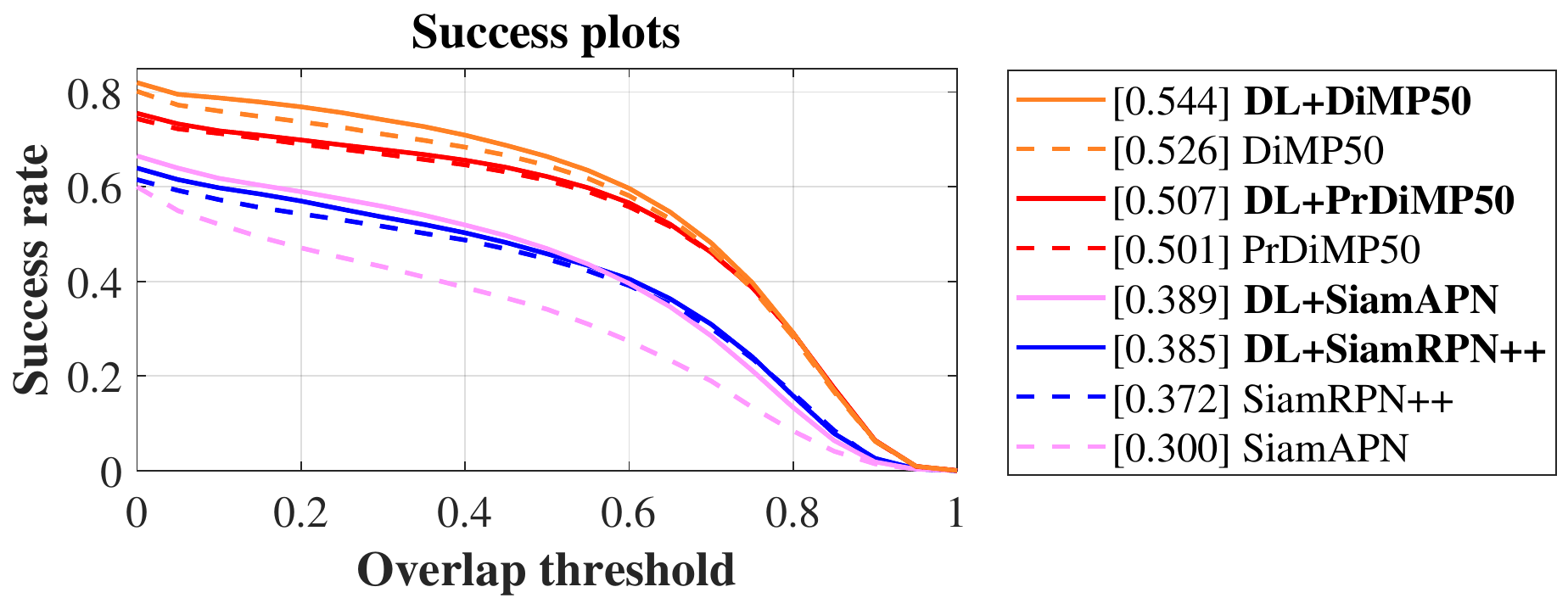}
	\includegraphics[width=0.49\linewidth]{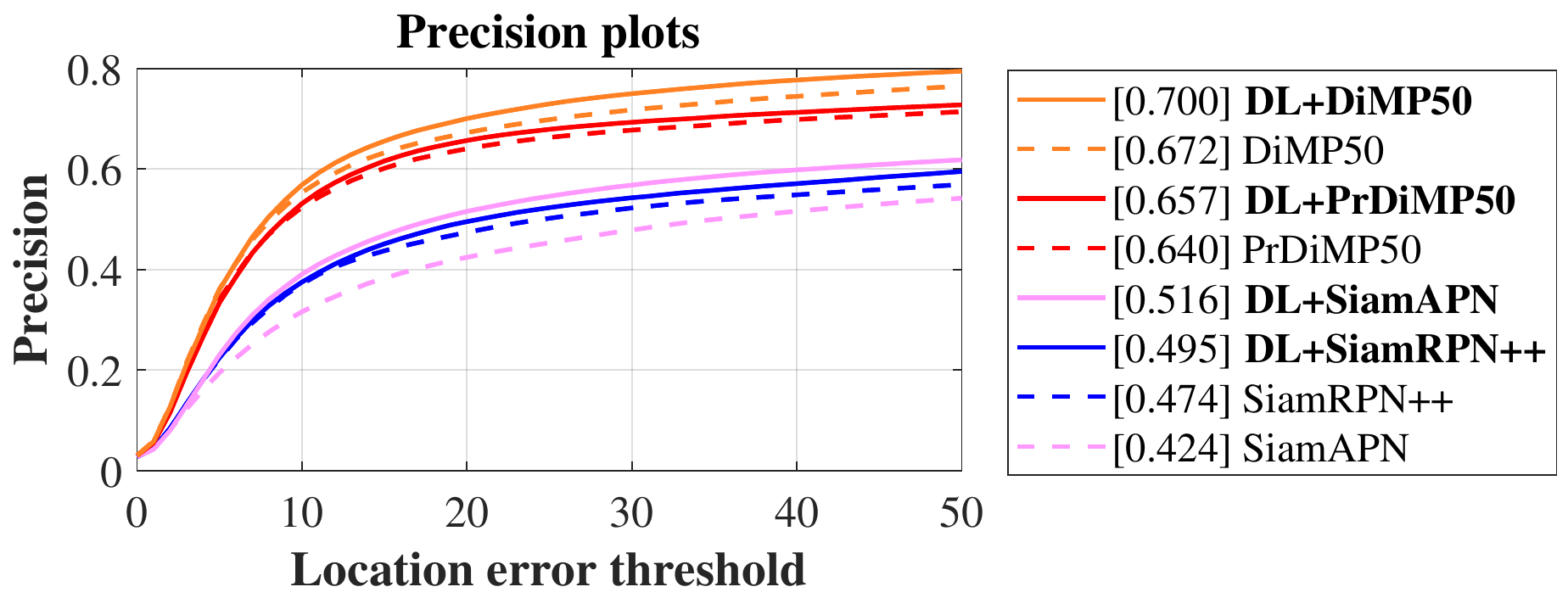}
	\setlength{\abovecaptionskip}{-4pt} 
	
	\caption
	{
		Precision and success plots of SOTA trackers with DarkLighter (denoted as DL in legend) enabled or not. DarkLighter promotes the performance of all involved trackers.
	}
	\label{fig:all}
\end{figure*}
\begin{table}[!b]
	\centering
	\caption{Gains after DarkLighter is implemented. }
	
	\resizebox{0.98\linewidth}{!}{
		\begin{tabular}{ccccc}
			\toprule
			Trackers & SiamAPN & SiamRPN++ & DiMP50 & PrDiMP50 \\
			\midrule
			Gains (\%) & +\textbf{29.67}/+\textbf{21.70} & +3.49/+4.43 & +3.42/+4.17 & +1.20/+2.66\\
			\bottomrule
	\end{tabular}}
	\label{tab:all}%
\end{table}%
\begin{table*}[!t]
	\setlength{\abovecaptionskip}{-0.1cm}
	\centering
	\caption{Attribute-based evaluation, data in the table are formulated as success/precision. $\Delta$ represents the percentages of DarkLighter boosted performance. DarkLighter raises tracking performance in typical challenges of dark tracking.}
	\renewcommand\tabcolsep{3pt}
	\resizebox{0.95\linewidth}{!}{
		\begin{tabular}{c|ccc|ccc|ccc|ccc}
			\hline
			\hline
			Trackers & \multicolumn{3}{c|}{DiMP50} & \multicolumn{3}{c|}{PrDiMP50} & \multicolumn{3}{c|}{SiamRPN++} & \multicolumn{3}{c}{SiamAPN} \\
			\hline
			Attributes & IV     & FM     & LR     & IV     & FM     & LR     & IV     & FM     & LR     & IV     & FM     & LR \\
			\hline
			Original & 0.502/0.641 & 0.510/0.642 & 0.504/0.704 & 0.462/0.588 & 0.483/0.614 & 0.454/0.625 & 0.330/0.426 & 0.361/0.449 & 0.347/0.487 & 0.279/0.400 & 0.283/0.400 & 0.289/0.449 \\
			+DarkLighter & 0.534/0.684 & 0.535/0.683 & 0.526/0.734 & 0.496/0.641 & 0.484/0.625 & 0.491/0.677 & 0.357/0.461 & 0.373/0.470 & 0.370/0.526 & 0.365/0.486 & 0.375/0.483 & 0.366/0.531 \\
			$\Delta$ (\%) & 6.37/6.71 & 4.90/6.39 & 4.37/4.26 & 7.36/9.01 & 0.21/1.79 & 8.15/8.32 & 8.18/8.22 & 3.32/4.68 & 6.63/8.01 & 30.82/21.5 & 32.51/20.75 & 26.64/18.26 \\
			\hline
			\hline
	\end{tabular}}
	\label{tab:att}%
\end{table*}%

\Remark All parameters and models of the approaches involved in the experiment remain their official releases without any tuning. 

\subsection{Implementation Details}
DarkLighter is implemented with PyTorch on a PC with an NVIDIA TITAN RTX GPU, a 32GB RAM, and an Intel i9-9920X CPU. NVIDIA Jetson AGX Xavier is employed as the real-world test platform. Images from the Part1 of SICE dataset~\cite{Cai2018TIP} are employed to train ME-Net. The weights $\lambda_1$, $\lambda_2$, $\lambda_3$, $\lambda_4$ and $\lambda_5$ of each loss function are set to 1600, 50, 10, 0.001, and 50, respectively. In the training process, images are resized to $256\times256$. The ADAM optimizer with a fixed learning rate of 0.0001 is adopted to optimize ME-Net. The batch size is set to 32, with a total of 193 epochs. The source code and some demo videos of DarkLighter are available at \url{https://github.com/vision4robotics/DarkLighter}.

\Remark To further alleviate the impact of various illumination conditions, the dataset consists of multi-exposure images is employed to train ME-Net, which further ensuring the generalization performance of DarkLighter in variation illuminations and promoting the ability of trackers to meet illumination variation challenges. 
\subsection{Evaluation Metrics}
Since the aim of this work is to facilitate UAV tracking aided by a low-light enhancer, the metrics of the high-level vision task, \textit{i.e.}, object tracking, are adopted to evaluate the experimental results. Following the one-pass evaluation (OPE)~\cite{Mueller2016ECCV}, precision and success rate are involved in the metrics. The precision is determined by the center location error (CLE) between the tracking result and the ground truth. The percentage of frames with a lower CLE than a given threshold is presented as the precision plot (PP), where the precision at 20 pixels is commonly used to rank trackers. As for success rate, intersection over union (IoU) between the estimated bounding box and the ground truth bounding box is adopted to measure the success rate. The percentage of frames that have larger IoU than the preset maximum threshold is reported as the success plot (SP). As common, the area-under-the-curve (AUC) on SP is used to rank the success rate of trackers.

\subsection{Efficacy of DarkLighter for UAV Dark Tracking}
\subsubsection{DarkLighter on Different Trackers}
As illustrated in Fig.~\ref{fig:all}, DarkLighter boosts the dark tracking performance clearly, with all trackers gain promotion in both precision and success rate. Specifically, the gains DarkLighter brings to each tracker are reported in TABLE~\ref{tab:all}. With an increase of more than \textbf{29\%} in AUC and \textbf{21\%} in precision, DarkLighter upgrades the dark tracking ability of SimpAPN significantly. Besides, DiMP50 ranks first place and retrieves its superior performance as in normal-light scenes to a certain extent with the help of DarkLighter. One can note that DarkLighter promotes trackers with different backbones, which demonstrates its generalization.

To evaluate the tracking performance of different challenges in detail, sequences in UAVDark135 are annotated with different attributes. Tracking performance in typical challenges of dark tracking, \textit{i.e.}, illumination variation (IV), fast motion (FM), and low-resolution (LR), is reported in TABLE~\ref{tab:att}. To be specific, in the term of IV, all involved trackers obtain promotions of over 6\%, which is credited to that DarkLighter restrains the effect of poor illumination greatly. In addition, thanks to the brightness brought by DarkLighter, trackers' ability to handle FM and LR are revived considerably. 

\begin{figure}[!t]
	\centering	
	\includegraphics[width=0.98\linewidth]{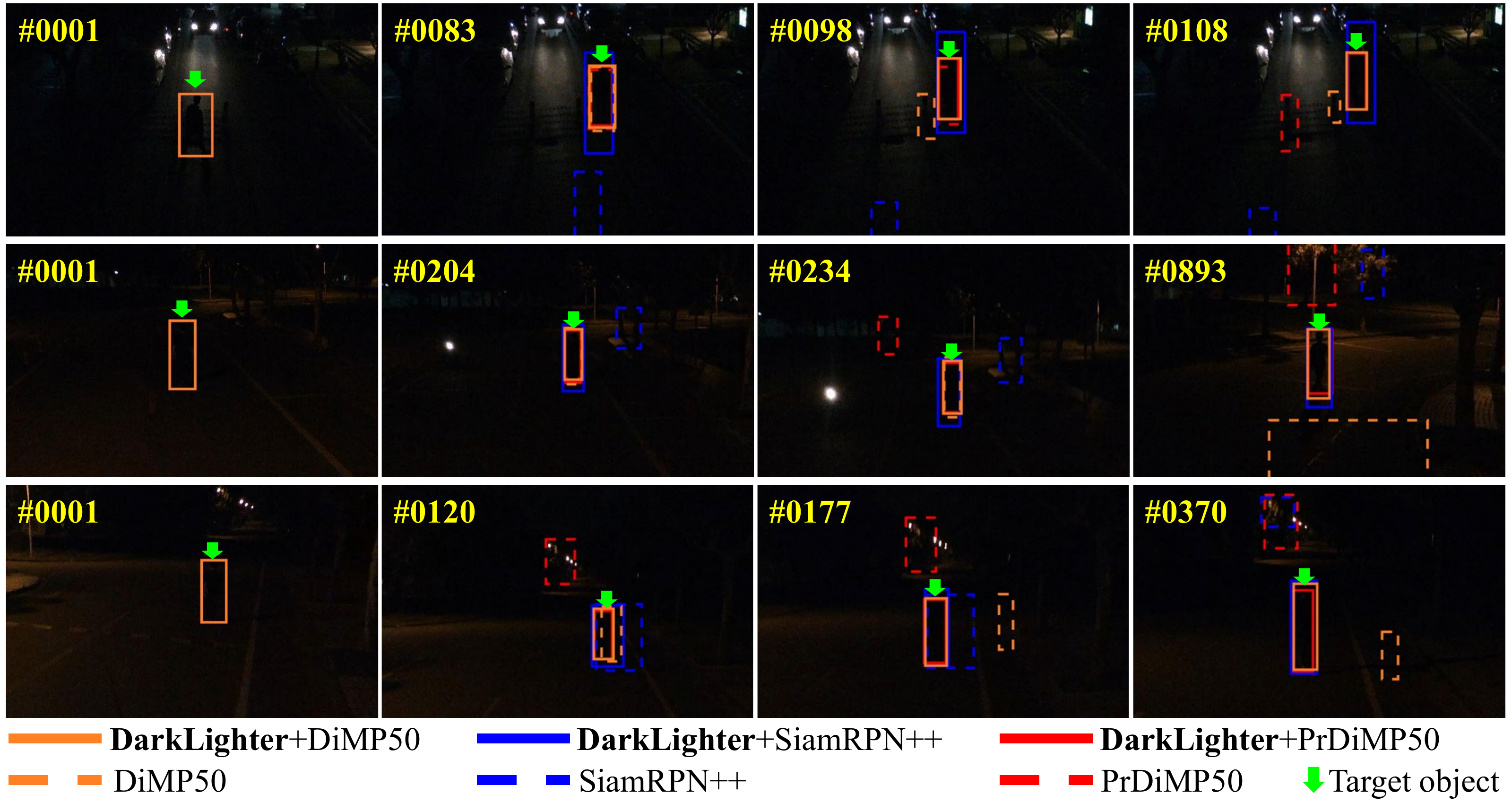}
	\setlength{\abovecaptionskip}{-2.5pt} 
	
	\caption
	{
		Some qualitative evaluation of trackers with DarkLighter enabled (solid line boxes) or not (dashed line boxes). From top to down, the sequences are $pedestrian3$, $person12\_2$, and $person12\_3$ from UAVDark135. Without the help of DarkLighter, the involved trackers fail to maintain robust tracking in these low-light conditions.
	}
	\label{fig:qua}
\end{figure}

Figure~\ref{fig:qua} shows some tracking screenshots of the trackers with DarkLighter enabled or not, DarkLighter raises the tracking reliability of the trackers in these low-light scenes.

\subsubsection{Comparison of DarkLighter and SOTA Enhancers}
\begin{figure}[!t]	
	\centering
	\includegraphics[width=0.97\linewidth]{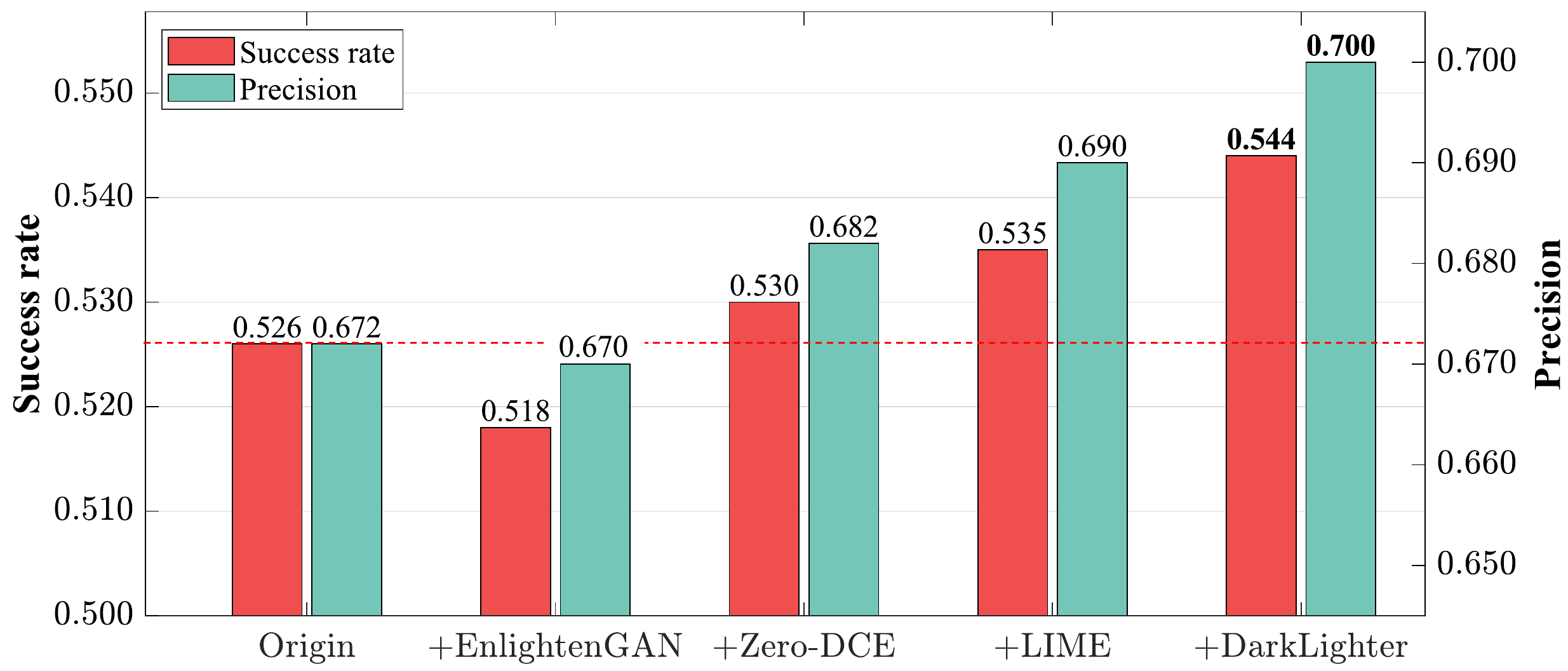}
	\setlength{\abovecaptionskip}{-2.5pt} 
	\caption
	{
		Comparison of DarkLighter and SOTA enhancers. DarkLighter brings more remarkable gains to object tracking.
	}
	\label{fig:comEN}
\end{figure}

To corroborate the advantages of DarkLighter for UAV dark tracking, the effect of DarkLighter and other SOTA low-light enhancers---EnlightenGAN~\cite{EnlightenGAN}, Zero-DCE~\cite{Guo_2020_CVPR}, and LIME~\cite{LIME}, on dark tracking is analyzed. Performance of tracking with images enhanced by different enhancers is exhibited in Fig.~\ref{fig:comEN}. Surprisingly, though EnlightenGAN improves the visual quality of images indeed, it causes the degradation of the tracking performance slightly. We conjecture that well visual quality can not guarantee satisfactory tracking accuracy. As for Zero-DCE and LIME, they elevate the tracking performance to a certain extent. Designed for robust dark tracking, DarkLighter arises tracking accuracy further, surpassing the second-best LIME by 1.68\% and 1.45\% in success rate and precision, respectively. 

\Remark Computational cost is put into great consideration in the design of DarkLighter. The proposed approach is constructed to be embedded system-friendly in terms of computational complexity. Though the baseline ranks the second place with the help of LIME, the introduced computational burn is too high to realize real-time tracking.

\begin{figure}[!t]	
\centering
\setlength{\abovecaptionskip}{-3pt} 
	\includegraphics[width=0.49\linewidth]{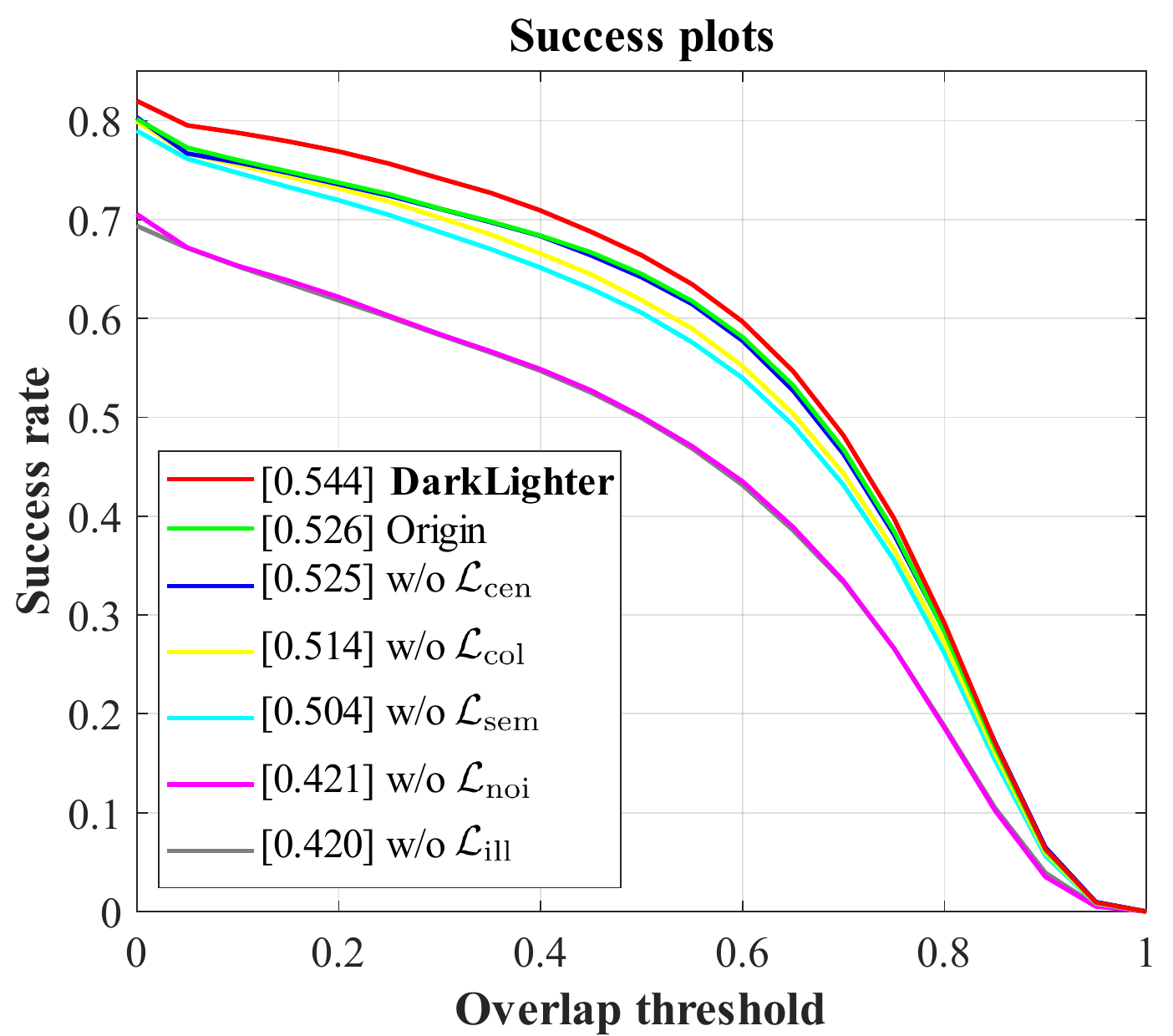}
	\includegraphics[width=0.49\linewidth]{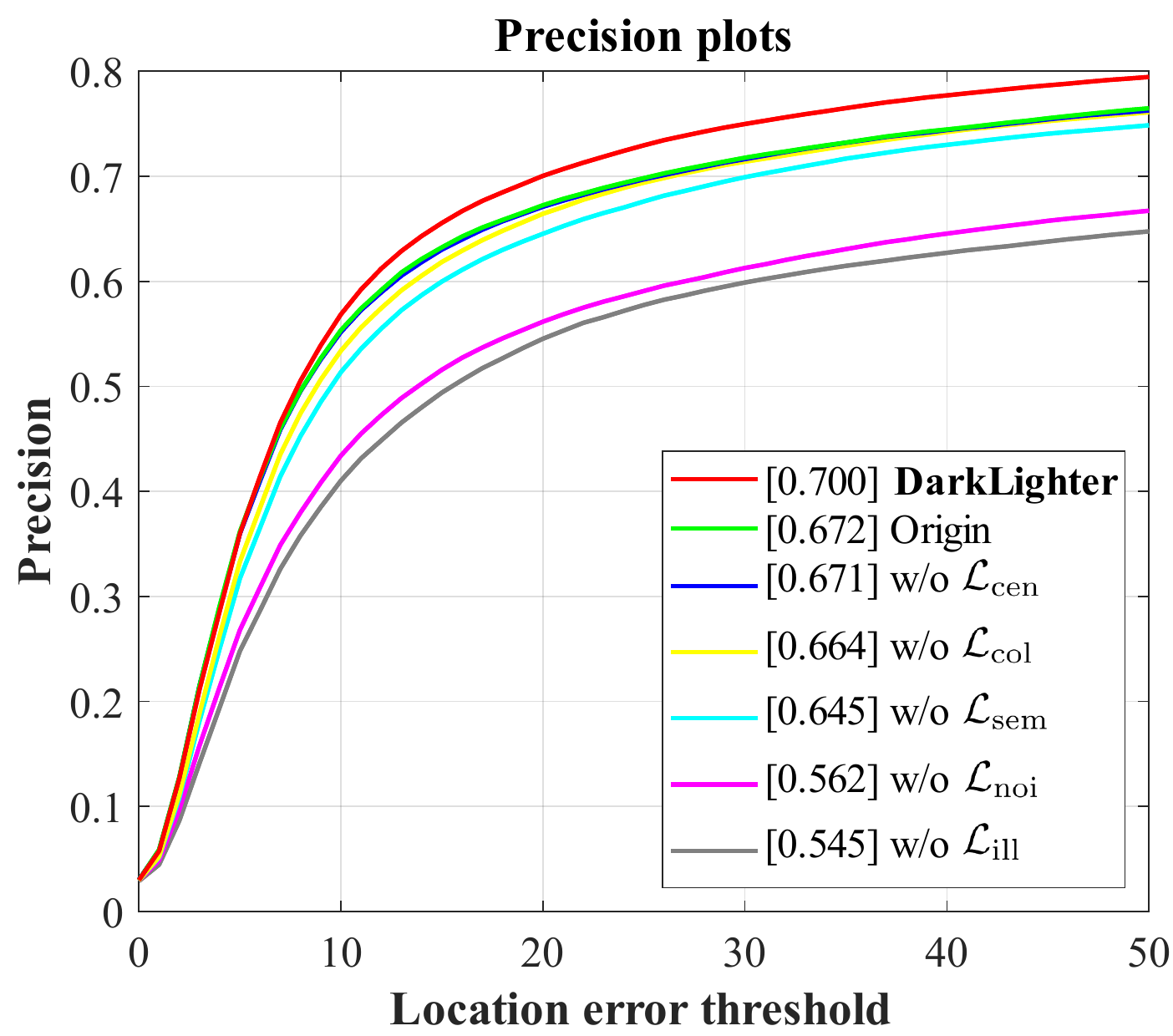}
	\setlength{\belowcaptionskip}{-2.5pt}
	 
	\caption
	{
		Contribution of each loss. $\mathcal{L}_{\rm col}$, $\mathcal{L}_{\rm cen}$, $\mathcal{L}_{\rm ill}$, $\mathcal{L}_{\rm sem}$, and $\mathcal{L}_{\rm noi}$ denote the color intensity loss, center-focused lightness loss, illumination regularization loss, semantic fidelity loss, and noise estimation loss, respectively.
	}
	\label{fig:abla}
\end{figure}

\subsection{Ablation Study}
The contribution of each loss in DarkLighter is further discussed in this subsection. 
Tracking performance of baseline under the effect of DarkLighter with different combinations of losses is illustrated in Fig.~\ref{fig:abla}. Without the illumination regularization loss $\mathcal{L}_{\rm noi}$ or the noise estimation loss $\mathcal{L}_{\rm noi}$, tracking performance degraded significantly, since $\mathcal{L}_{\rm ill}$ and $\mathcal{L}_{\rm noi}$ guide the estimation of illumination and noise maps directly. Semantic fidelity loss $\mathcal{L}_{\rm sem}$ preserves the semantic information of the enhanced image from degradation, tracking robustness cannot maintain without it. Removing the color intensity loss $\mathcal{L}_{\rm col}$, generated unrealistic images can influence the feature extraction of the backbone, leading to tracking failure. Moreover, removing the center-focused lightness loss, the light condition of the object is not satisfying for tracking, thus can hardly guide robust tracking. With the activation of all loss functions, the potential of DarkLighter is fully exploited, arising baseline's success rate (0.526) and precision (0.672) to 0.544 and 0.700, respectively. To sum up, all loss functions in DarkLighter are pivotal and contribute to the facilitating of tracking performance.
\begin{figure}[!t]
	\setlength{\abovecaptionskip}{-4pt} 
	
	\centering	
	\includegraphics[width=0.99\linewidth]{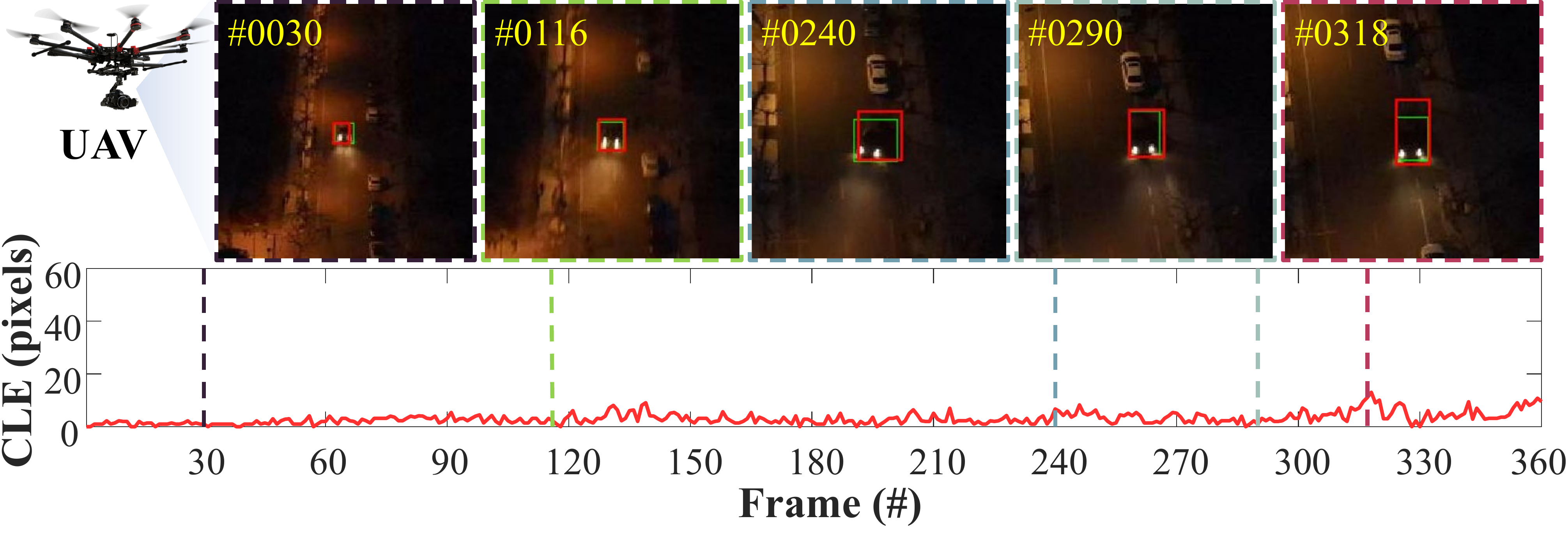}
	
	\includegraphics[width=0.99\linewidth]{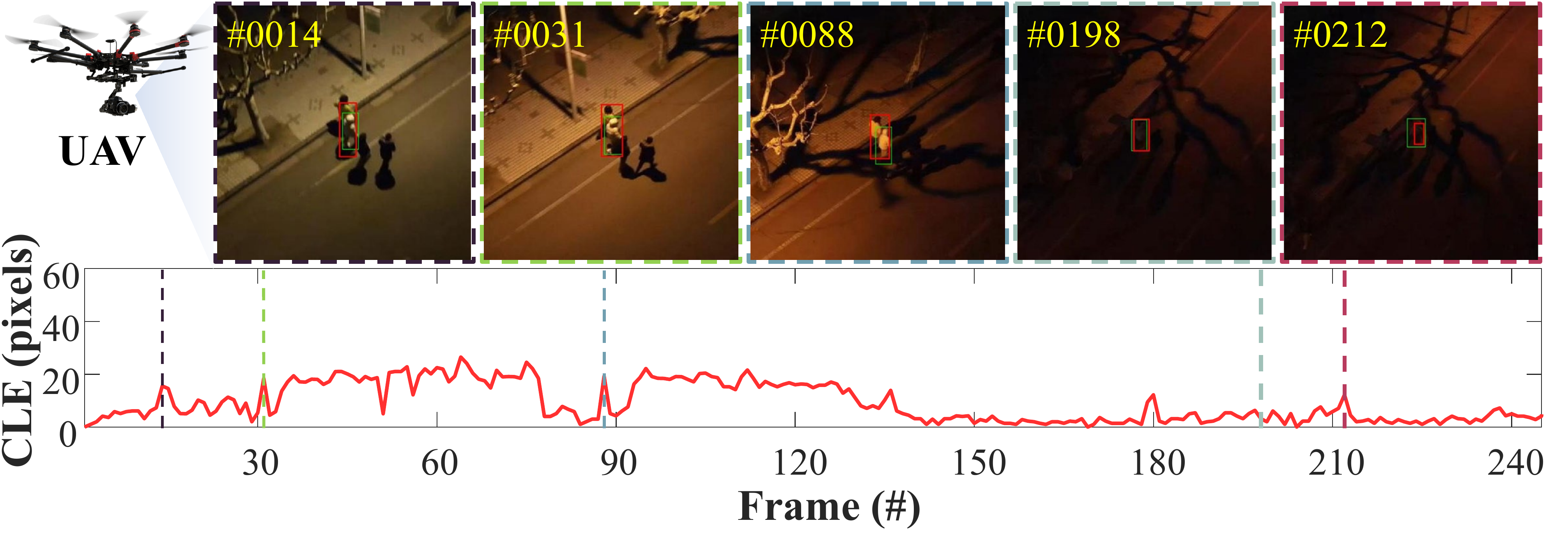}
	
	\includegraphics[width=0.99\linewidth]{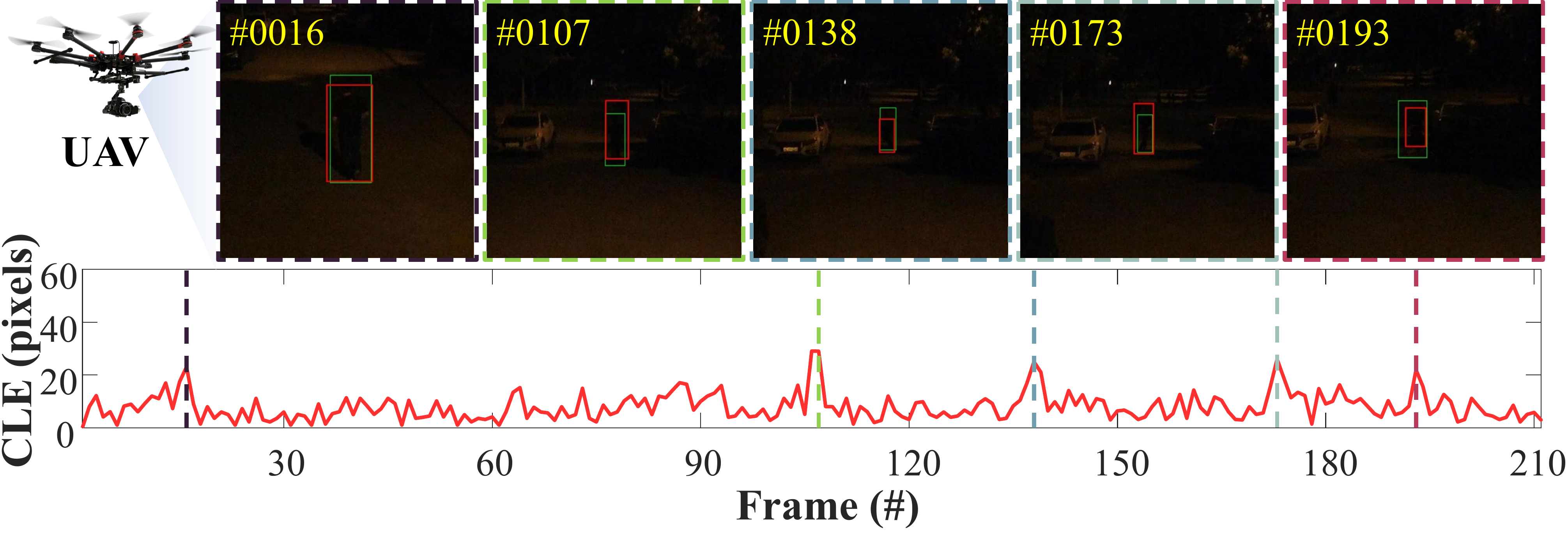}
	
	\setlength{\belowcaptionskip}{-8pt} 
	\caption
	{
		Real-world tests onboard a typical UAV platform. \NoOne{Red} and \NoTwo{green} bounding boxes denote tracking results and ground truth, respectively. CLE curves between estimated locations and ground truth bounding boxes are reported below the tracking snapshots. The baseline tracker yields robust performance in low-light conditions with the activation of DarkLighter.
	}
	\label{fig:real}
\end{figure}

\begin{table}[!b]
	\setlength{\abovecaptionskip}{-0.15cm}
	
	\setlength{\belowcaptionskip}{-2cm}
	\centering
	\caption{Average millisecond per frame (MSPF), frame per second (FPS), parameters (Params), and floating-point operations (FLOPs) of DarkLighter.}
	\resizebox{0.59\linewidth}{!}{
		\begin{tabular}{cccc}
			\toprule
			MSPF   & FPS    & Params (K) & FLOPs (G) \\
			\midrule
			9.20   & 108.70 & 74.77  & 4.90 \\
			\bottomrule
	\end{tabular}}
	\label{tab:model}%
	
\end{table}%

\subsection{Real-World Tests}
To corroborate the practicability and dependability of DarkLighter for UAV tracking, it is implemented onboard a typical UAV platform. TABLE~\ref{tab:model} reports the complexity analysis of the proposed approach. What stands out is the high inference speed ($>$100 FPS) and lightweight ($<$75K parameters) of DarkLighter, making it applicable for UAV. Further, the top-ranked tracker in evaluation---DiMP50, is adopted to tracking in darkness with the activation of DarkLighter. Figure~\ref{fig:real} presents some screenshots and CLE curves of three real-world tests. 
Specifically, the main challenges in test~1 contain low-light, illumination variation, and small object, \textit{etc}. Due to the pretreatment of DarkLighter for original patches, the backbone can extract discriminative features, helping distinguish the small object from the darkness. As for test 2, there are similar objects around and the object undergoes drastic changes in appearance caused by light condition variation. In test 3, the object runs far and is swallowed by the darkness. Although there are some sudden tracking biases in the tracking process, the baseline recalls the object and maintains robust tracking with the assistant of DarkLighter. 
In summary, the real-world tests indicate that DarkLighter is efficient and dependable for raising tracking performance in low-illumination.

\section{Conclusion}
This work constructs a general low-light enhancer towards facilitating high-level vision tasks, exemplified by UAV tracking. To this concern, DarkLighter dedicates to alleviate the influence of poor illumination on UAV tracking. Experiments verify its compatibility and effectiveness on different tracking approaches. Comparison with current SOTA low-light enhancers demonstrates the superiority of DarkLighter for promoting UAV tracking robustness in low-illumination conditions. Real-world tests onboard a typical UAV system further substantiate its applicability and dependability, with low computational consumption. To sum up, we strongly believe this study can be of assistance to broadening UAV tracking applications to low-light conditions.

\section*{Acknowledgment}

This work is supported by the National Natural Science Foundation of China (No. 61806148) and the Natural Science Foundation of Shanghai (No. 20ZR1460100).


\bibliographystyle{IEEEtran}
\normalem
\bibliography{IROS2021}

\end{document}